\documentclass[sigconf]{acmart}

\usepackage{amsmath}
\usepackage{amsfonts}

\copyrightyear{2026}
\acmYear{2026}
\setcopyright{cc}
\setcctype{by}
\acmConference[SA Technical Communications '26]{SIGGRAPH Asia 2026 Technical Communications}{December 01--04, 2026}{Kuala Lumpur, Malaysia}
\acmBooktitle{SIGGRAPH Asia 2026 Technical Communications (SA Technical Communications '26), December 01--04, 2026, Kuala Lumpur, Malaysia}
\acmDOI{10.1145/3829339.3847833}
\acmISBN{979-8-4007-2841-9/2026/12}

\begin{document}

\title{Gaussian Image Steganography via Parameter-Domain Keyed Embeddings}

\author{Tong Wu}
\orcid{0009-0004-9865-7103}
\authornote{Corresponding author.}
\affiliation{%
  \institution{East China University of Science and Technology}
  \city{Shanghai}
  \country{China}}
\email{24012920@mail.ecust.edu.cn}

\author{Runze Cheng}
\orcid{0009-0000-2629-9635}
\authornote{Also with Computational Light Laboratory, University College London.}
\affiliation{%
  \institution{University of Cambridge}
  \city{Cambridge}
  \country{United Kingdom}}
\email{rc2062@cam.ac.uk}

\author{Xiaoyue Fan}
\orcid{0009-0003-3216-0121}
\affiliation{%
  \institution{University College London}
  \city{London}
  \country{United Kingdom}}
\email{merryxyfan@gmail.com}

\author{Kaan Akşit}
\orcid{0000-0002-5934-5500}
\affiliation{%
  \institution{University College London}
  \city{London}
  \country{United Kingdom}}
\email{kaanaksit@kaanaksit.com}

\renewcommand{\shortauthors}{Wu et al.}

\begin{abstract}
2D Gaussian-based image representation is becoming increasingly popular, and our work proposes a new approach to steganography by embedding information within Gaussian parameters rather than image pixels. We first fit the parameters of this representation to the target image and employ a secret key to select a subset of these parameters for fine-tuning, allowing us to embed an 8-bit message while maintaining high visual fidelity in the reconstructed image. Thirty fitting experiments on three synthetic images show that the correct key can recover the message without error, while decoding with incorrect keys yields a Bit Error Rate (BER) of \(0.543\), close to random guessing. Compared with random selection using the key, selecting the least-disturbing edits recovers the message more reliably (one-sided \(p=0.031\)), and the average PSNR cost is only \(0.091\) dB in visual quality. The embedding method transfers to 112 natural images at \(256\times256\) using 4,096 Gaussians. The correct-key BER is \(0.000\), and decoding under a wrong key stays close to random guessing at \(0.520\). Our method embeds the payload through three Gaussian parameter types: log-anisotropy, opacity, and color luminance. In a separate nine-fit reduced setting, color luminance is removed, so the payload uses two instead of three parameter types, a \(33.3\%\) reduction; all 256 Gaussians remain in the fitted representation. The correct key still recovers the message without error. However, wrong-key BER rises from \(0.514\) to \(0.571\), moving farther from random guessing (\(0.5\)).
\end{abstract}


\maketitle

\section{Introduction}
\label{sec:introduction}

A 2D Gaussian-based image representation models an image as a set of 2D Gaussian primitives~\citep{Zhang2024GaussianImage}. Meanwhile, people often hide additional information, also known as a payload, in images for purposes such as watermarking. The conventional practice is to modify the pixel values of a given image in an imperceptible way. However, whether information can be stored in the parameters of image-space 2D Gaussian primitives and recovered from the modified parameters remains underexplored.

Image-space neural watermarking and latent-space steganography methods hide the payload in pixels or latent representations, and can be decoded from the reconstructed image~\citep{Zhu2018HiDDeN,Tancik2020StegaStamp,Bui2023RoSteALS}. 3D-GSW embeds watermarks in 3D Gaussian scenes and recovers them from rendered images~\citep{Jang2025_3DGSW}. GaussianMarker extracts messages from both Gaussian parameters and rendered images~\citep{Huang2024GaussianMarker}. Our payload is read from the fitted image-space 2D Gaussian parameter set, whereas the rendered image alone is insufficient for decoding. This parameter requirement adds an extra layer of security because an image-only copy does not provide the modified Gaussian parameters required by the decoder. If one refits a new set of Gaussian parameters from the rendered image, the hidden information is not retained. We therefore focus on sharing the parameter set itself.

Our method starts by identifying Gaussians to hide the secret information with the help of a secret key. Among the Gaussians that can be safely modified, the key generates a repeatable set of candidates and maps their parameter changes to the message bits. We then select the least-disturbing combination that encodes the message. We recover the embedded information from those Gaussians using the same secret key. In our tests, the tested wrong keys yield a mean BER close to 0.5. We evaluate our method against two baselines: (1) \textbf{random selection with the key}; and (2) \textbf{selection of the least-disturbing edits without the key}.    Notably, we find that reducing the available parameter types preserves correct-key recovery but moves wrong-key decoding farther from random guessing. We have released our code at \href{https://github.com/tongwu-research/gaussian-image-steganography}{GitHub: \nolinkurl{tongwu-research/gaussian-image-steganography}}.

\begin{figure*}[t]
  \centering
  \includegraphics[width=0.80\textwidth]{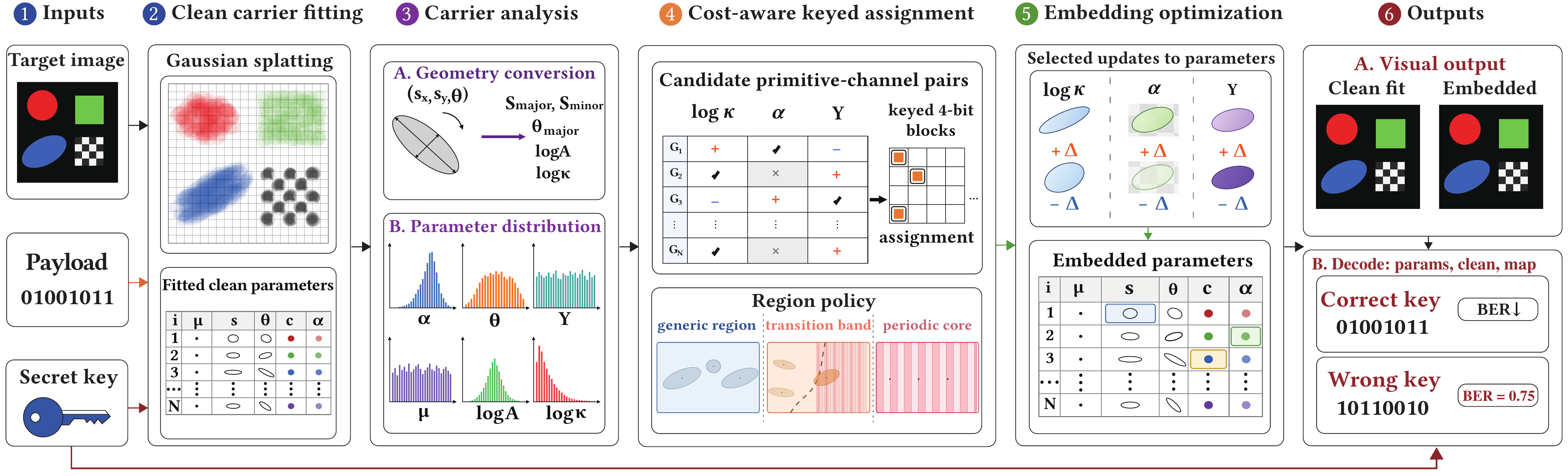}
  \caption{\textbf{System overview.} A clean 2D Gaussian carrier is fitted from an input image. Canonical carrier channels are analyzed and scored. Keyed carrier assignment selects signed parameter updates, and decoding compares correct-key recovery against wrong-key behavior.}
  \Description{A full-width system diagram showing the pipeline from inputs through clean carrier fitting, canonical carrier analysis, cost-aware keyed assignment, embedding optimization, and key-specific decoding.}
  \label{fig:system_overview}
\end{figure*}

\section{Method}
\label{sec:method}

\paragraph{2D Gaussian Image Representation.} We work with a bounded image-space formulation built from 2D Gaussian primitives~\citep{Zhang2024GaussianImage}, following the pipeline in Figure~\ref{fig:system_overview}. This differs from 3D Gaussian Splatting (3DGS), which represents 3D scenes with 3D Gaussian primitives~\citep{Kerbl2023_3DGS}.
We represent an image with \(N\) primitives \(G_i\). Each primitive \(G_i\) is parameterized by a center offset \(\boldsymbol{\mu}_i \in \mathbb{R}^2\), a scale vector \(\mathbf{s}_i = (s_{x,i}, s_{y,i})^T \in \mathbb{R}^2\), an in-plane rotation angle \(\theta_i \in \mathbb{R}\), an RGB color vector \(\mathbf{c}_i \in [0,1]^3\), and an opacity \(\alpha_i \in [0,1]\). Let \(P=\{(\boldsymbol{\mu}_i,s_{x,i},s_{y,i},\theta_i,\mathbf{c}_i,\alpha_i)\}_{i=1}^{N}\) denote the parameter set of all primitives. We render the image on the normalized 2D image grid \(\mathbf{x} = (x, y)\) by accumulated blending,
\begin{equation}
\hat{\mathbf{I}}(\mathbf{x}) = \sum_{i=1}^{N} \alpha_i \, G_i(\mathbf{x}) \, \mathbf{c}_i ,
\end{equation}
where
\[
\begin{gathered}
G_i(\mathbf{x}) = \exp\!\bigl(-\tfrac{1}{2}(\mathbf{x}-\boldsymbol{\mu}_i)^T \Sigma_i^{-1}(\mathbf{x}-\boldsymbol{\mu}_i)\bigr),\\
\Sigma_i = R(\theta_i)\,\mathrm{diag}(s_{x,i}^2,\,s_{y,i}^2)\,R(\theta_i)^T,
\end{gathered}
\]
with \(R(\theta_i)\) the in-plane rotation matrix. There is no transmittance, depth ordering, or multiview consistency in this bounded 2D setting, and our method acts directly on the parameters of \(G_i\) to hide the desired payload.

\begin{figure}[tbp]
  \centering
  \includegraphics[width=0.78\columnwidth]{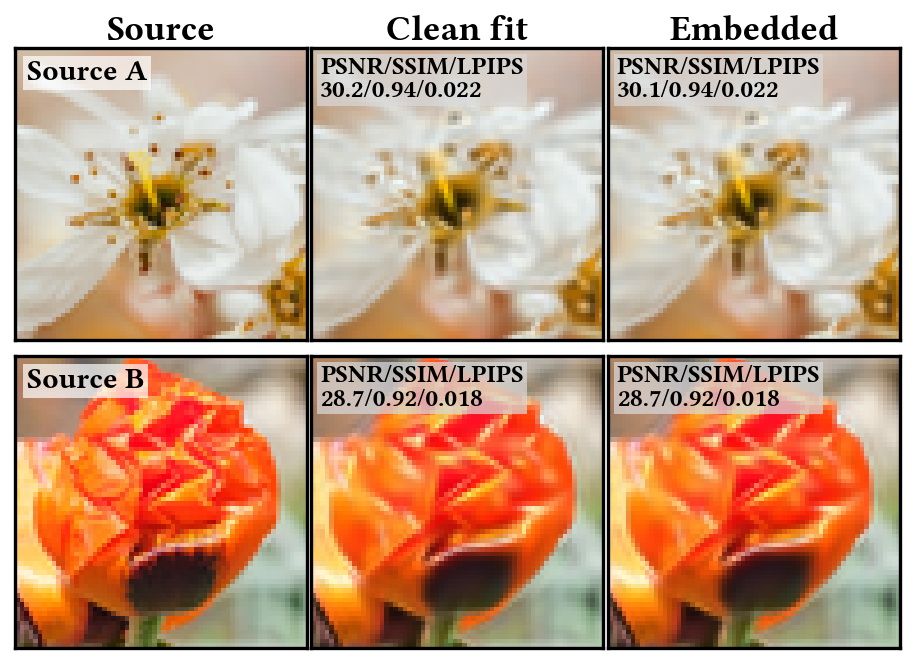}
  \caption{\textbf{Clean Fit vs. Embedded Reconstruction.} We refer to the clean fit as the reconstruction of a source image from our pipeline without payload embedding, whereas the embedded case additionally applies the keyed embedding pipeline. Images are \(64\times64\) pixels. Source photographs are by \href{https://commons.wikimedia.org/wiki/File:White_flowers_in_close-up_(Unsplash).jpg}{Christian Widell} and \href{https://commons.wikimedia.org/wiki/File:Poppy_flower_overview.jpg}{Christian J. Leuner}, respectively.}
  \Description{A single-column fitting overview figure comparing two source images, their clean fits, and their embedded renderings.}
  \label{fig:fitting_overview}
\end{figure}

\paragraph{Clean Image Fitting.} First, we fit the target image by optimizing \(P\) over \(\{G_i\}_{i=1}^{N}\). On the normalized grid, we obtain \(\widehat{P}\) by minimizing
\begin{equation}
\begin{aligned}
\widehat{P}
&\leftarrow \arg\min_{P} \mathcal{L}\!\left(\hat{\mathbf{I}}(\cdot;P), \mathbf{I}^\star\right),\\
\mathcal{L}\!\left(\hat{\mathbf{I}}(\cdot;P), \mathbf{I}^\star\right)
&= \frac{1}{M}
\sum_{m=1}^{M}
\left\|
\hat{\mathbf{I}}(\mathbf{x}_m;P) - \mathbf{I}^\star(\mathbf{x}_m)
\right\|_2^2 .
\end{aligned}
\end{equation}
Here, \(\hat{\mathbf{I}}(\mathbf{x}_m;P)\) denotes the image rendered from \(P\) at sampled pixel \(\mathbf{x}_m\), \(m \in \{1,\dots,M\}\), and \(\mathbf{I}^\star\) is the target image. Figure~\ref{fig:fitting_overview} shows the source images, clean fits, and embedded reconstructions.

\paragraph{Canonical Parameterization and Carrier Channels.} We utilize a subset of Gaussian parameters as the carrier. Following the scale--rotation covariance parameterization used in Gaussian splatting \citep{Kerbl2023_3DGS,Huang2024_2DGS}, carrier design uses a canonical parameter space instead of the raw \((s_{x,i},s_{y,i},\theta_i)\) representation. Let \(s_{\mathrm{major},i}=\max(s_{x,i},s_{y,i})\) and \(s_{\mathrm{minor},i}=\min(s_{x,i},s_{y,i})\).
Swapping the axis ordering rotates the angle by \(\pi/2\) and wraps it into \([0,\pi)\), giving a canonical major-axis angle \(\theta_{\mathrm{major},i}\). From these we derive two scale summaries:
\begin{equation}
\log A_i = \log\!\big(s_{\mathrm{major},i} s_{\mathrm{minor},i}\big),
\qquad
\log \kappa_i = \log\!\left(\frac{s_{\mathrm{major},i}}{s_{\mathrm{minor},i}}\right),
\end{equation}
giving log-area and log-anisotropy. This removes the axis-ordering ambiguity and makes scales directly comparable. Guided by the fitted distributions in Supplementary Figure~S2, we use three carrier channels for the payload: log-anisotropy, opacity, and color luminance \(Y_i\) (the luminance of \(\mathbf{c}_i\)). We exclude rotation because angular edits are less stable; log-area and position are diagnostics.

\paragraph{Parameter-Domain Embedding Formulation.} Given a payload \(\mathbf{b} \in \{0,1\}^B\) and a secret key \(k\), we modify the fitted clean parameter set \(\widehat P\) into \(\widetilde P\). When decoding with \(k\), the payload can be decoded correctly, while decoding under other keys should remain near chance on average.  In our experiments, \(\mathbf{b}\) is an 8-bit ASCII character, though the formulation handles any bit vector. Unlike existing 3DGS algorithms, our pipeline neither adds nor removes any \(G_i\) in the reconstruction process. We keep \(\alpha_i\) and \(\mathbf{c}_i\) independent to avoid removing a payload channel.

\paragraph{Cost-Aware Carrier Scoring.} For every signed action \((i,\ell,\pm)\), where \(\ell\) indexes the candidate channel, we define four cost terms: rendering sensitivity \(\sigma_{i,\ell}\), distribution shift \(d_{i,\ell}^{\pm}\), visual importance \(v_{i,\ell}\), and outlier-value penalty \(t_{i,\ell}\). Their weighted sum is \(\rho_{i,\ell}^{\pm}=w_s\sigma_{i,\ell}+w_d d_{i,\ell}^{\pm}+w_v v_{i,\ell}+w_t t_{i,\ell}\). A sensitivity analysis of these weights is provided in Supplementary Section~S7.

\paragraph{Structure-Aware Region Policy for Periodic Images.} Since periodic structures are visually sensitive to parameter perturbations, we partition the fitted Gaussians into three regions with different carrier permissions and perturbation strengths: the ``periodic core'', ``transition band'', and ``generic region''. We estimate a dominant spatial frequency from the source and determine the stripe-normal direction and its orthogonal tangent. For each Gaussian we measure three components contributing to the periodicity score: repetition intensity \(r_i\), direction alignment \(a_i\), and normalized edge support \(e_i\). Their product defines the periodicity score \(\phi_i=r_i a_i e_i\). See Supplementary Section~S2 for details. \begingroup
\begin{itemize}
\item ``Periodic core'' --- values above the 75th percentile. It retains only \(Y_i\), with a scale factor of \(0.5\).
\item ``Transition band'' --- values above the 50th percentile, after excluding core members; lower-percentile primitives can also enter when edge and repetition strength are high. It uses scale factors of \(0.5\) for \(\alpha_i\) and \(0.75\) for \(Y_i\).
\item ``Generic region'' --- the rest. These use the payload-channel set \((\log\kappa_i,\alpha_i,Y_i)\) with a perturbation scale factor of \(1.0\).
\end{itemize}
\endgroup
\noindent We select these percentile cutoffs empirically in preliminary experiments. They remain fixed in all reported runs and may not be optimal for every image. Not all eligible \(G_i\) are safe carriers: we exclude rotation entirely, and further filter out candidate actions located in high-edge, high-importance, or stripe-aligned periodic positions before assignment. We organize the remaining carrier actions into keyed 4-bit blocks: the key determines the initial candidate signed actions and the block parity mapping from signed actions to payload bits. Block assignment then follows a syndrome-inspired method~\citep{FillerJudasFridrich2011STC}: signed actions are selected according to the region-dependent channel rules and coverage balancing constraints, which limit the concentration of selected actions. Concretely, the assignment minimizes the total action cost \(\sum_{i,\ell}\rho_{i,\ell}^{\pm}\) subject to the key-induced parity constraints. A signal-to-noise score then refines the less reliable block assignments. In periodic sources, the cost avoids luminance edits where they are most visible, preferring scale (\(\log\kappa_i\)) and opacity (\(\alpha_i\)) edits.

\paragraph{Embedding Optimization and Regularization.} Once a keyed assignment is in place, we fine-tune selected carriers to embed the payload while keeping image and parameter distribution close to the clean fit. We minimize
\begin{equation}
\begin{aligned}
\mathcal{L}_{\mathrm{embed}} ={}&
  \lambda_{\mathrm{msg}}\,\mathcal{L}_{\mathrm{msg}}
  + \lambda_{\mathrm{fid}}\,\mathcal{L}_{\mathrm{fid}} \\
&{}+ \lambda_{\mathrm{imp}}\,\mathcal{L}_{\mathrm{imp}}
  + \lambda_{\mathrm{key}}\,\mathcal{L}_{\mathrm{key}}
  + \lambda_{\mathrm{aux}}\,\mathcal{L}_{\mathrm{aux}}.
\end{aligned}
\end{equation}
The five terms enforce message recovery, rendered-image fidelity, parameter-distribution similarity, key specificity, and auxiliary robustness, respectively. See Supplementary Section~S3 for details.

\paragraph{Decoding and Wrong-Key Evaluation.} Decoding uses the embedded parameters \(\widetilde P\), the corresponding clean reference parameters \(\widehat P\), and an assignment map produced by the encoder under key \(k\). The map specifies which parameter changes to read and how to convert them into message bits. Decoding reads the realized action scores from the deviation \(\widetilde P - \widehat P\), thresholds them into per-action binary decisions, and applies the block parity mapping to recover \(\widehat{\mathbf{b}}_k\). See Supplementary Section~S2 for the extraction rule.
The primary metric is bit error rate,
\begingroup
\begin{equation}
\operatorname{BER}(\widehat{\mathbf{b}}_k,\mathbf{b})
=
\frac{1}{B}
\sum_{j=1}^{B}
\mathbf{1}\!\left[\hat{b}_{k,j} \neq b_j\right].
\end{equation}
\endgroup
Correct-key BER checks whether the payload survives embedding. Wrong-key BER evaluates decoding under assignments generated from unrelated keys. Wrong-key BER near \(0.5\) indicates key specificity, with larger deviations indicating weaker specificity.

\section{Implementation}
\label{sec:implementation}

\paragraph{Training.} We first optimize a clean Gaussian fit and then fine-tune the selected carrier parameters for payload embedding. Optimization settings accompany the code.

\paragraph{Data.} We evaluate three \(3\times64\times64\) RGB targets---a flat gradient, a periodic stripe pattern, and a checkerboard pattern---each fit with 10 random seeds and 256 Gaussian primitives (see Supplementary Figure~S1).  For the \(64\times64\), 256-primitive setting, the 8-bit payload corresponds to \(0.031\) bits per primitive, or \(0.00195\) bits per rendered pixel. The payload is carried by the selected carrier actions.

\section{Evaluation}
\label{sec:experiments}

\paragraph{Setup.}  The synthetic targets are a flat gradient, a periodic stripe, and a checkerboard (see Supplementary Figure~S1). Synthetic targets isolate flat, periodic, and mixed structure. For each synthetic target, we independently fit 256 Gaussians using 10 random seeds that were not used during method development, giving 30 fitted Gaussian sets in total. We evaluate every method on the same 30 sets. Under a fixed \(k\), each set receives an 8-bit payload. We report three metrics: correct-key \(\operatorname{BER}_{\mathrm{ck}}\!\downarrow\), mean wrong-key \(\operatorname{BER}_{\mathrm{wk}}\) over 16 wrong keys (target \(0.5\)), and \(\Delta\mathrm{PSNR}\) from the clean fit. Both optimization and checkpoint selection use the same 16 wrong keys.

\paragraph{Comparison on Synthetic Targets.}

Table~\ref{tab:main_results} compares our method with two baselines. Both use the same allowed Gaussian parameter edits, 8-bit payload, and spatial restrictions as our method. \emph{Matched-random} uses the key but chooses edits at random. \emph{Greedy no-key} chooses the lowest-cost allowed edits without a key. We do not compare against radiance-field or 3DGS watermarking methods, which evaluate a different 3D setting. Our method and greedy no-key recover all 240 payload bits. Matched-random makes payload errors on 5 of the 30 sets and matches our method on the other 25. The paired permutation \(p\)-values are \(0.031\) (one-sided) and \(0.062\) (two-sided). Mean wrong-key BER is \(0.543\) for our method and \(0.565\) for matched-random. Our wrong-key BER stays between \(0.4\) and \(0.6\) on 27 of the 30 sets. Matched-random has a smaller \(\Delta\mathrm{PSNR}\), but it does not recover every payload. Most \(\Delta\mathrm{PSNR}\) values for our method are near \(0\), while a few outliers raise the mean. Filtering leaves 182--422 eligible parameter edits per set, but only 2--5 Gaussians are modified.  

\begin{table}[!tbp]
\caption{Results on 30 fitted Gaussian sets with an 8-bit payload. Values are reported as mean \(\pm\) standard deviation.}
\centering
\small
\setlength{\tabcolsep}{3.5pt}
\begin{tabular}{l ccc}
\toprule
Method & \(\operatorname{BER}_{\mathrm{ck}}\!\downarrow\) & \(\operatorname{BER}_{\mathrm{wk}}\!\to\!0.5\) & \(\Delta\mathrm{PSNR}\!\downarrow\) \\
\midrule
Matched random          & \(0.042_{\pm 0.111}\) & \(0.565_{\pm 0.066}\) & \(0.066_{\pm 0.129}\) \\
Greedy no-key           & \(\mathbf{0.000}_{\pm 0.000}\) & N/A               & \(0.075_{\pm 0.206}\) \\
Ours                    & \(\mathbf{0.000}_{\pm 0.000}\) & \(\mathbf{0.543}_{\pm 0.059}\) & \(0.091_{\pm 0.154}\) \\
\bottomrule
\end{tabular}

\label{tab:main_results}
\end{table}

\begin{table}[!tbp]
\caption{Ablation results on a separate nine-fit development set. All rows use the same nine fits and 8-bit payload. Values are reported as mean \(\pm\) standard deviation.}
\centering
\small
\setlength{\tabcolsep}{3.5pt}
\begin{tabular}{l ccc}
\toprule
Ablation & \(\operatorname{BER}_{\mathrm{ck}}\!\downarrow\) & \(\operatorname{BER}_{\mathrm{wk}}\!\to\!0.5\) & \(\Delta\mathrm{PSNR}\!\downarrow\) \\
\midrule
Ours                  & \(\mathbf{0.000}_{\pm 0.000}\) & \(\mathbf{0.514}_{\pm 0.026}\) & \(0.584_{\pm 1.348}\) \\
\(-\) Sinkhorn        & \(0.028_{\pm 0.055}\) & \(0.512_{\pm 0.048}\) & \(0.509_{\pm 0.846}\) \\
2-channel only        & \(\mathbf{0.000}_{\pm 0.000}\) & \(0.571_{\pm 0.057}\) & \(1.021_{\pm 1.869}\) \\
\(-\) region policy   & \(0.069_{\pm 0.110}\) & \(0.567_{\pm 0.054}\) & \(0.011_{\pm 0.024}\) \\
\(-\) coverage balancing & \(0.042_{\pm 0.125}\) & \(0.523_{\pm 0.039}\) & \(0.899_{\pm 1.496}\) \\
\bottomrule
\end{tabular}

\label{tab:ablation}
\end{table}

\paragraph{Ablation Study.} Table~\ref{tab:ablation} removes one component at a time on the nine-fit development set. Without the Sinkhorn distribution term~\citep{Cuturi2013Sinkhorn}, some correct-key errors appear, but wrong-key BER remains near \(0.5\). The two-channel variant uses only log-anisotropy and opacity. It recovers every payload bit, but its wrong-key BER increases to \(0.571\). The most periodic region permits only luminance edits, so this region cannot carry the payload in the two-channel variant. Removing the region policy or coverage balancing, which spreads edits across image regions, also introduces correct-key errors and moves wrong-key BER away from \(0.5\). The small \(\Delta\mathrm{PSNR}\) without the region policy is not a comparable improvement because that variant does not recover every bit.  

\paragraph{Robustness.}  Using the seed-0 fit of each synthetic target, we test all 256 possible 8-bit byte values. The average correct-key BER is \(0.030\), while wrong-key decoding stays close to chance (\(0.500\)). In total, \(623/768\) target-byte cases decode correctly. Errors concentrate on the periodic stripe target, the hardest case.  Supplementary Section~S6 and Figure~S3 report recovery and \(\Delta\mathrm{PSNR}\) at different payload lengths on the nine-fit development set. We then add noise to the payload-carrying parameters of nine fitted sets. The uniform and Gaussian noise have standard deviations of approximately \(0.0058\) and \(0.005\), respectively. Supplementary Table~S1 reports four noise scales over 27 evaluations each. Mean correct-key BER rises to \(0.042\) under uniform noise and \(0.060\) under Gaussian noise, while wrong-key BER stays near \(0.5\). After 200 additional optimization steps using only the reconstruction loss, correct-key BER rises to \(0.264\). A new Gaussian fit of the rendered image raises correct-key BER to about \(0.75\), so recovery fails.

\paragraph{Natural Images.} We use 24 Kodak images~\citep{FranzenKodak} and 100 DIV2K validation crops~\citep{Agustsson2017DIV2K} prepared at \(3\times256\times256\) to embed the fixed 8-bit payload.  All methods succeed with 8,192 Gaussians, so we reduce the count. We select 4,096 Gaussians using a separate set of 12 images and evaluate all three methods on the remaining 112 images. Our method and greedy no-key recover all 112 payloads, while matched-random recovers 110. Our wrong-key BER remains close to chance at \(0.520\). Mean clean-fit PSNR is \(33.68\) dB, with \(97/112\) images above \(30\) dB. On the same covers, pretrained RoSteALS recovers 109 of 112 bytes at \(26.77\) dB, while our method recovers all 112 at \(57.24\) dB (Supplementary Table~S2). RoSteALS needs no key for decoding.

\begin{table}[!tbp]
\caption{Results on 112 held-out RGB images at \(3\times256\times256\) using 4,096 Gaussians and a fixed 8-bit payload. Recovery requires all eight bits to be correct. Values are mean \(\pm\) standard deviation.}
\centering
\small
\setlength{\tabcolsep}{3.5pt}
\begin{tabular}{l c ccc}
\toprule
Method & \shortstack{Payloads\\recovered \(\uparrow\)} & \(\operatorname{BER}_{\mathrm{ck}}\!\downarrow\) & \(\operatorname{BER}_{\mathrm{wk}}\!\to\!0.5\) & \(\Delta\mathrm{PSNR}\!\downarrow\) \\
\midrule
Matched random & \(110/112\) & \(0.003_{\pm 0.026}\) & \(0.525_{\pm 0.019}\) & \(0.013_{\pm 0.023}\) \\
Greedy no-key  & \(\mathbf{112/112}\) & \(\mathbf{0.000}_{\pm 0.000}\) & N/A & \(0.006_{\pm 0.014}\) \\
Ours           & \(\mathbf{112/112}\) & \(\mathbf{0.000}_{\pm 0.000}\) & \(\mathbf{0.520}_{\pm 0.029}\) & \(0.009_{\pm 0.016}\) \\
\bottomrule
\end{tabular}

\label{tab:natural}
\end{table}

\paragraph{Conclusion.}

We introduce a novel steganography method for image representations based on 2D Gaussians, embedding payloads directly in the Gaussian primitives rather than image pixels. Experiments on natural images and controlled synthetic cases---flat, periodic, and mixed structures---show that the correct key reliably recovers the payload, while other keys yield near-random decoding. Our method adds a new level of complexity in steganography methods, benefiting content creators with additional security against malicious activities.

\bibliographystyle{ACM-Reference-Format}
\bibliography{method_refs_20260409_compact}

\end{document}